\documentclass[twoside]{article}

\usepackage[accepted]{aistats2024_arxiv}

\usepackage[round]{natbib}

\usepackage[utf8]{inputenc} %
\usepackage[T1]{fontenc}    %
\usepackage{hyperref}       %
\usepackage{url}            %
\usepackage{booktabs}       %
\usepackage{amsfonts}       %
\usepackage{nicefrac}       %
\usepackage{microtype}      %
\usepackage{xcolor}         %
\usepackage[pdftex]{graphicx}
\usepackage{amsmath}
\usepackage{amssymb}
\usepackage{bm}
\usepackage{bbm}

\DeclareMathOperator{\Bern}{Bern}
\DeclareMathOperator{\Beta}{Beta}
\DeclareMathOperator{\LogNormal}{LogNormal}

\def\eqref#1{equation~\ref{#1}}

\def\1{\bm{1}}

\def\rva{{\mathbf{a}}}
\def\rvb{{\mathbf{b}}}

\def\rvh{{\mathbf{h}}}
\def\rvu{{\mathbf{i}}}

\def\rvm{{\mathbf{m}}}

\def\rvt{{\mathbf{t}}}
\def\rvu{{\mathbf{u}}}

\def\rvx{{\mathbf{x}}}

\def\rmW{{\mathbf{W}}}

\DeclareMathAlphabet{\mathsfit}{\encodingdefault}{\sfdefault}{m}{sl}
\SetMathAlphabet{\mathsfit}{bold}{\encodingdefault}{\sfdefault}{bx}{n}

\usepackage{cleveref}
\crefformat{equation}{#2(#1)#3}
\let\eqref\cref

\begin{document}

\runningauthor{Jake C. Snell, Gianluca Bencomo, Thomas L. Griffiths}

\twocolumn[

\aistatstitle{A Metalearned Neural Circuit for Nonparametric Bayesian Inference}

\aistatsauthor{
Jake C. Snell$^1$ \\  \And
Gianluca Bencomo$^1$ \And
Thomas L. Griffiths$^{1,2}$
}

\aistatsaddress{
$^1$Department of Computer Science \> $^2$Department of Psychology \\
Princeton University\\
\texttt{\{js2523,gb5435,tomg\}@princeton.edu}
} ]

\begin{abstract} 
 Most applications of machine learning to classification assume a closed set of balanced classes. This is at odds with the real world, where class occurrence statistics often follow a long-tailed power-law distribution and it is unlikely that all classes are seen in a single sample. Nonparametric Bayesian models naturally capture this phenomenon, but have significant practical barriers to widespread adoption, namely implementation complexity and computational inefficiency. To address this, we present a method for extracting the inductive bias from a nonparametric Bayesian model and transferring it to an artificial neural network. By simulating data with a nonparametric Bayesian prior, we can metalearn a sequence model that performs inference over an unlimited set of classes. After training, this ``neural circuit'' has distilled the corresponding inductive bias and can successfully perform sequential inference over an open set of classes. Our experimental results show that the metalearned neural circuit achieves comparable or better performance than particle filter-based methods for inference in these models while being faster and simpler to use than methods that explicitly incorporate Bayesian nonparametric inference.
\end{abstract}

\section{Introduction}

Standard machine learning approaches to classification assume that the set of possible classes is known \emph{a priori}. Classification in this setting thus involves selecting the most appropriate class label from a closed set.
However, this is not the case for human learners. Imagine European explorers in Australia seeing a kangaroo for the first time. Rather than trying to classify this observation into an existing class -- is it a deer or a rabbit? -- they recognized that a new class needs to be created. Although this is easy for humans, current machine learning systems struggle to identify novel classes and use them in predictions~\citep{parmar2023openworld}.

Bayesian statistics offers an elegant solution to the problem of novel classes: define a model in a way that does not make a commitment to an upper bound on the number of classes. This idea is expressed in nonparametric Bayesian models, namely the Dirichlet process mixture model (DPMM) \citep{antoniak1974mixtures}. In this model, a new data point is assumed to be generated from an existing class with probability proportional to the number of previous observations from that class and from a new class with a probability proportional to $\alpha$, a hyperparameter of the model. This makes it possible to both postulate that a new datapoint might come from a new class and capture long-tailed distributions of class frequency commonly found in real-world classification problems. 

Despite their elegance, nonparametric Bayesian models have fallen out of favor in machine learning, as they are difficult to reconcile with the current focus on large-scale models defined over complex objects such as images. The scalability of non-parametric Bayesian models is limited due to the high computational cost of Bayesian inference, typically requiring the use of sampling algorithms such as Markov chain Monte Carlo \citep{neal2000markov} or particle filters \citep{fearnhead2004particle}. Applying these models to complex objects requires creativity in defining generative models that are sufficiently expressive without making inference intractable. Furthermore, most standard Bayesian inference algorithms are not designed to perform sequential inference but rather assume that all data are presented together as a single batch.

\begin{figure*}[t!]
  \centering
    \includegraphics[width=450pt]{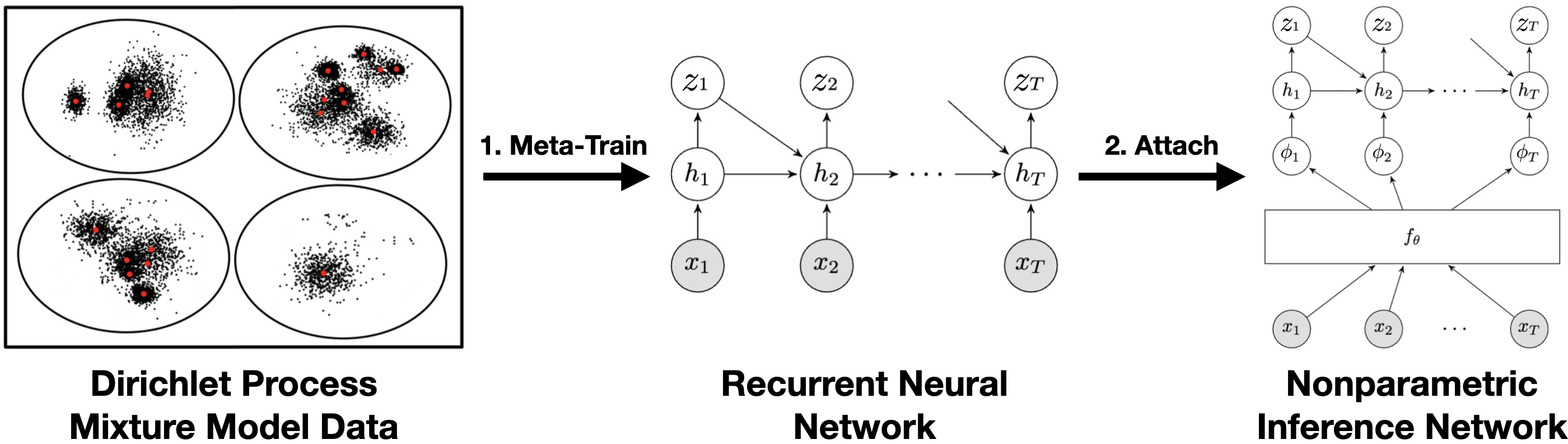}
    \caption{Our proposed nonparametric inference network first internalizes the desired nonparametric Bayesian prior via metalearning a recurrent neural network (RNN) to model its posterior distribution over class assignments. Afterwards, the metalearned RNN, or ``neural circuit,'' has captured the corresponding inductive bias and can be used to perform sequential inference over a potentially unbounded number of classes.}
    \label{fig:circuit}
\end{figure*}

In this paper we pursue a different approach to inference in nonparametric Bayesian models: training a recurrent neural network (RNN) to approximate the posterior distribution over classes from a DPMM. We formulate this problem as one of {\em metalearning}, %
repeatedly sampling a sequence of class memberships and observations from the DPMM and training the model to predict the class of each observation conditioned on the class labels of those preceding it. The resulting \emph{neural circuit} (Figure~\ref{fig:circuit}) internalizes the inductive bias of the nonparametric Bayesian model used to generate the training tasks.

Our neural circuit approach combines the elegance of Bayesian nonparametric models with the predictive power of deep learning in a principled and practical manner. The discriminative nature of the RNN allows it to successfully classify complex inputs without making the restrictive distributional assumptions necessitated by standard Bayesian inference algorithms. Since RNNs are fundamentally sequence models, the neural circuit is able to efficiently make predictions for a sequence of observations as each predictive distribution is computed in constant time. We apply our neural circuit to a challenging open-set image classification task and find that it achieves excellent predictive performance at a fraction of the computational cost of standard DPMM inference algorithms.

\section{Background}\label{sec:background}

To motivate the problem setup, consider the task of a learner classifying objects in a new environment. The learner is presented with a sequence of items $\rvx_1, \rvx_2, \ldots, \rvx_T$. After each item $\rvx_t$ is observed, the learner attempts to predict its class label $z_t$. The learner is then apprised of the true class label and subsequently must make a prediction for the class label of $\rvx_{t+1}$. In order to perform this task successfully across timesteps, the learner needs to successfully accomplish two subgoals: (a) identifying to which class an item belongs (possibly creating a new class if necessary), and (b) updating its internal representation of the corresponding class after learning the true class of an item. Expressed in probabilistic terms, the predictive distribution over the sequence of labels $z_{1:T} = z_1, \ldots, z_T$ can be written as
\begin{equation}
    p(z_{1:T} | \rvx_{1:T}) = p(z_1 | \rvx_1) \prod_{t=2}^T p(z_t | \rvx_{1:t}, z_{1:t-1}). \label{eq:sequential_predictive_distribution}
\end{equation}
We see in \eqref{eq:sequential_predictive_distribution} both of these subgoals: (a) is captured since the predictive distribution at a given timestep $p(z_t | \rvx_{1:t}, z_{1:t-1})$ is conditioned on the current item $\rvx_t$, and (b) is captured in the transition from timestep $t$ to $t+1$, since after the transition, both $\rvx_t$ and $z_t$ are now conditioned upon and able to influence future predictions. In the remainder of this section, we first review Dirichlet process mixture models, which provide a framework for expressing \eqref{eq:sequential_predictive_distribution} as posterior inference over $z_{1:T}$ without restricting each $z_t$ to belong to a closed set. We then review the particle filter of \citet{fearnhead2004particle}, which offers a method for sequentially computing the predictive distribution through the use of weighted particles, each representing a joint assignment of class labels.

\subsection{Dirichlet Process Mixture Models}

The nonparametric Bayesian solution to the prediction task presented above involves infinite mixture models, the most notable of which is the Dirichlet process mixture model (DPMM)~\citep{antoniak1974mixtures,escobarw95,rasmussen1999infinite}. There are two components to the DPMM: a distribution over class memberships and a class-conditional distribution over observations. In this model observations are generated from latent classes, where the distribution over classes places no limit on the number of classes. The joint probability of a sequence of observations and class labels follows a Markov process on $z_{1:T}$ and assumes class-conditional independence for $\rvx_{1:T}$:
\begin{equation}
    p(z_{1:T}, \rvx_{1:T}) = p(z_1) p(\rvx_1 | z_1) \prod_{t=2}^T p(z_t | z_{1:t-1}) p(\rvx_t | z_t). \label{eq:dpmm_joint_prob}
\end{equation}
The prior distribution on class memberships is relatively simple.
The conditional distribution $p(z_t|z_{1:t-1})$ is
\begin{equation}
    p(z_t = k \mid z_{1:t-1}) \propto \left \{ 
    \begin{array}{cl}
    n_k & \mbox{$k$ previously observed} \\
    \alpha & \mbox{$k$ is a new class},
    \end{array}
    \right . \label{eq:crp_definition}
\end{equation}
where $n_k$ denotes the number of occurrences of class $k$ in $z_1, \ldots, z_{t-1}$, and by convention the value of $k$ for a new class is taken to be one higher than the number of classes observed so far. This process is known as the Chinese restaurant process (CRP)~\citep{aldous1985exchangeability}. 

The conditional distribution over observations is
\begin{equation}
p(\rvx_t \mid z_t = k) = g(\rvx_t | \bm{\varphi}_k),
\end{equation}
where $g( \cdot | \bm{\varphi})$ is some probability function with parameter $\bm{\varphi}$. Each $\bm{\varphi}_k$ is in turn distributed according to a shared prior $\pi(\bm{\varphi}_k)$ for $k = 1, 2, \ldots$. Producing the predictive distribution \eqref{eq:sequential_predictive_distribution} is a simple application of Bayes' rule to the joint distribution \eqref{eq:dpmm_joint_prob}. 

Direct computation of the posterior in a DPMM is intractable, and therefore several methods have been developed to perform approximate inference using methods such as MCMC~\citep{neal2000markov} and variational inference~\citep{blei2006variational}. However, these methods typically assume that all observations are presented simultaneously (i.e.\ the batch setting) and do not attempt to handle the sequential nature inherent to our problem formulation.

\subsection{Particle Filter for the DPMM}\label{sec:background_particle_filter}

One notable method that \emph{does} aim to perform sequential inference of class labels in the DPMM is the particle filter proposed by~\citet{fearnhead2004particle}. Particle filters~\citep{doucet2009tutorial,chopin2020introduction} maintain a set of weighted particles at each timestep that approximate the posterior distribution over latent variables. At a high level, a particle filter propagates a weighted set of particles forward in time by first sampling a new state for each particle from a transition distribution and then assigning each sample a new weight according to a potential function. If the transition distribution and potential function are chosen such that the product of the transition probability and potential are proportional to the target posterior, then the new set of weighted particles remains a valid approximation of the posterior at the next timestep. This sampling step is followed by an optional resampling step that may randomly drop low weight particles and replace them with additional copies of high weight particles. The resampling step sacrifices some of the quality of the current timestep's approximation with the hope that extra copies of high weight particles will lead to better quality estimates of the posterior at future timesteps.

 In order to make inference tractable in the DPMM setting, \citet{fearnhead2004particle} assumes that $g(\rvx | \bm{\varphi})$ belongs to the exponential family and $\pi(\bm{\varphi})$ is its corresponding conjugate prior. In this case, it is possible to marginalize over $\bm{\varphi}$ when computing the posterior predictive distribution for a class. Suppose that $\rvx_i \sim g(\rvx_i | \bm{\varphi})$ for $i = 1, \ldots, n$. Then there exists a closed form expression for the posterior predictive
\begin{equation}
    p(\rvx_{n+1} | \rvx_{1:n}) = \int p(\rvx_{n+1} | \bm{\varphi}) p(\bm{\varphi} | \rvx_{1:n}) \, d\bm{\varphi},\label{eq:expfamily_posterior_predictive}
\end{equation}
since the posterior $p(\bm{\varphi} | x_{1:n})$ has the same form as the conjugate prior $\pi(\bm{\varphi})$.

Each particle in the method of \citet{fearnhead2004particle} represents an entire trajectory of class labels up until the current timestep. Suppose that after $t$ timesteps there are $J$ particles $z_{1:t}^{(j)}$ each with weight $w_j \ge 0$ for $j = 1, \ldots, J$ such that $\sum_{j=1}^J w_j = 1$. For each particle, the transition distribution is defined as:
\begin{align}
p(z_{t+1} = k \mid \rvx_{1:t+1}, z_{1:t}^{(j)}) \propto& \, p(z_{t+1} =k | z_{1:t}^{(j)})  \\ &p(\rvx_{t+1} | \rvx_{1:t}, z_{1:t}^{(j)}, z_{t+1} = k), \nonumber
\end{align}
where the likelihood term $p(x_{t+1} | z_{1:t}^{(j)}, z_{t+1} = k)$ can be computed in closed form according to \eqref{eq:expfamily_posterior_predictive}. In the case that $k$ represents a new class, the corresponding term will be the prior predictive probability instead.

Notice that the particle filter accomplishes both of the subgoals put forth at the beginning of this section. For a new observation $x_{t+1}$, the particle filter assigns probability to each class label proportional to how well the class explains the observation as measured by the class's posterior predictive distribution (prior predictive in the case of a new label). Additionally, after each true class label is revealed, the corresponding class's representation is updated since all future posterior predictive distributions computed for the class will be conditioned on the corresponding observation. This may be practically implemented by maintaining a set of running sufficient statistics for each class.

Two major shortcomings of the particle filter are that it makes restrictive distributional assumptions (the class conditional distribution needs to be exponential family) and requires many particles in order to sufficiently approximate the posterior. In the next section, we present our metalearned neural circuit which is aimed at remedying these shortcomings.

\section{Metalearning a Neural Circuit}

We propose a novel amortized inference approach for inference in the DPMM, based on using metalearning to train a recurrent neural network (RNN) to predict class memberships. By applying metalearning to tasks defined by sampling data from a DPMM, we can create a model that emulates Bayesian inference in this model. After training, the RNN has internalized the corresponding inductive bias, hence the name ``neural circuit.''

Our approach is inspired by the observation that the updates of the particle filter in Section~\ref{sec:background_particle_filter} can be implemented by accumulating sufficient statistics of observations. Recall that an exponential family distribution can be expressed in terms of natural parameters $\bm{\eta}$:
\begin{equation}
    p(\rvx | \bm{\eta}) = h(\rvx) \exp \left\{ \langle \bm{\eta}, \mathbf{t}(\rvx) \rangle - A(\bm{\eta}) \right\},
\end{equation}
where $\mathbf{t}(\rvx)$ are the sufficient statistics and $A(\bm{\eta})$ is the log normalizer. The corresponding conjugate prior is
\begin{equation}
    p(\bm{\eta} | \bm{\tau}, \nu) = \exp \left\{ \langle \bm{\tau}, \bm{\eta} \rangle - \nu A(\bm{\eta}) - B(\bm{\tau}, \nu) \right\}.
\end{equation}
The posterior after observing $\rvx_{1:n}$ is of the same form as the prior, namely $p(\bm{\eta} | \bm{\tau}', \nu')$ where
\begin{align}
    \bm{\tau}' &= \bm{\tau} + \sum_{i=1}^n \mathbf{t}(\rvx_i) \\ 
    \nu' &= \nu + n.
\end{align}
From this perspective, the particle filter can be implemented by first initializing the representation of each class to be $\bm{\tau}, \nu$. Then the sufficient statistics $\mathbf{t}(\rvx_t)$ for each observation are extracted and used to update the corresponding class's $\bm{\tau}$ after the true label is revealed.

A loosely analogous computation is carried out by a recurrent network, which given some input $\rvx_t$ and previous hidden state $\rvh_{t-1}$, computes an updated hidden representation $\rvh_t$ and an output $\rvu_t$:
\begin{equation}
    \rvh_t, \rvu_t \leftarrow \textsc{RNN}_{\bm{\theta}}(\rvx_t, \rvh_{t-1})
\end{equation}
The differences with respect to the particle filter are twofold: the representation of each cluster is no longer separate but shared in $\rvh_t$, and the representations and updating procedure are end-to-end learnable.

We also recognize that both computational subgoals introduced in Section~\ref{sec:background} can be accomplished by the RNN: the output $\rvu_t$ can be the basis for predicting the current class label, and $\rvh_t$ can be made to capture the updated current state of all classes simultaneously. More specifically, we define the predictive distribution to be
\begin{align}
    p_{\bm{\gamma}}(z_t | \rvx_{1:t}, z_{1:t-1}) &= \textsc{Softmax}(\rva_t + \rvm_t)  \\
    \rva_t &= \rmW \rvu_t + \rvb \nonumber \\
    m_{tk} &=  \left\{ \begin{array}{cl}
    0 & k \le 1 + \max z_{1:t-1} \\
    -\infty & \mbox{otherwise},
    \end{array}
    \right . \nonumber \\
    \rvu_t, \rvh_t &\leftarrow \textsc{RNN}_{\bm{\theta}}([\rvx_t, \textsc{One-Hot}(z_{t-1})], \rvh_{t-1}), \nonumber
\end{align}
where $\bm{\gamma} \triangleq \{ \bm{\theta}, \rmW, \rvb\}$ are learnable parameters. To predict the current class label, the RNN cell's output $\rvu_t$ is mapped to a provisional logit $\rva_t$ using a learnable weight matrix $\rmW$ and bias vector $\rvb$. The logits are then additively masked by $\rvm_t$, which preserves the logit as long as predicting the corresponding class would be valid (i.e.\ the class label is at most one greater than any previously seen class label). The input to the RNN is the concatenation of the current observation $\rvx_t$ and a one-hot representation of the previous label $z_{t-1}$ (all zeros when $t=1$).

In order to learn the weights of this ``neural circuit,'' we turn to metalearning. In metalearning, a system is presented with a set of tasks sampled from a distribution over tasks. The goal is to  leverage the shared structure of these tasks not only to become better at solving each individual task but also to solve future tasks better, effectively ``learning to learn'' \citep{schmidhuber1987evolutionary,bengio1992optimization,baxter97a,hochreiter2001learning}. This is done by estimating a set of hyperparameters shared across tasks. In our case, these are the weights of the neural circuit, which we metalearn by minimizing negative log-likelihood of item-label sequences generated by a DPMM. Let $z_{1:T}$ be generated according to a CRP \eqref{eq:crp_definition} and $\rvx_t \mid z_t = k$ be generated according to some (possibly unknown) distribution $g(\rvx_t | \bm{\varphi}_k)$ for $t = 1, \ldots, T$. Let $\mathcal{D}$ denote this joint distribution over $(z_{1:T}, \rvx_{1:T})$. We define
\begin{align}
    \bm{\gamma}^* = \arg\min_{\bm{\gamma}} \mathbb{E}_{(z_{1:T}, \rvx_{1:T}) \sim \mathcal{D}}\left[ \mathcal{L}(z_{1:T}, \rvx_{1:T} ; \bm{\gamma}) \right] \\
    \mathcal{L}(z_{1:T}, \rvx_{1:T} ; \bm{\gamma} )=  -\frac{1}{T} \sum_{t=1}^T \log p_{\bm{\gamma}}(z_t | \rvx_{1:t}, z_{1:t-1}),
\end{align}
where the minimization is performed via a gradient-based optimization procedure. This metalearning approach can be applied equally well whether the the form of the class-conditional distributions is known or not, since all that is required are samples from the joint distribution. If the class-conditional distribution is unknown, the distribution over $\rvx_t$ is thus taken to be the empirical distribution constructed by placing an evenly-weighted point mass on each data sample within a particular class.

In addition to potentially reducing the computational cost, training an RNN to perform inference in a nonparametric Bayesian model results in a differentiable neural circuit that can potentially be integrated into other models. We explore a limited version of this in our experiments (Section~\ref{sec:experiments}), in which the neural circuit is applied on top of activations extracted from a pretrained convolutional neural network.

\paragraph{Relationship to Amortized Inference} Our method can be viewed as a form of amortized inference, wherein a function (e.g.\ represented by a deep neural network) is learned to directly map from inputs to an approximate posterior distribution over latent variables~\citep{kingma2014auto,rezende2014stochastic}. Amortized inference is similar to discriminative classification in that it directly maps from inputs to class labels, but it can also be viewed as approximate Bayesian inference within the generative modeling framework. Networks trained to perform amortized inference are sensitive to the choice of prior, since different priors will lead to different posteriors, the KL-divergence to which will be minimized during learning. In a similar fashion, our method is also affected by the choice of DPMM prior as different priors change the distribution over sequence that the neural circuit is trained on.

\paragraph{Relationship to Metalearning} A variety of methods have been used to learn a shared set of hyperparameters to solve a set of related tasks problem for deep neural networks \citep{vinyals2016matching, santoro2016meta,andrychowicz2016learning,ravi2017optimization,finn2017model}, but our approach is most similar to those in which a recurrent neural network is trained to produce appropriate actions across tasks \citep{duan2016rl,wang2016learning}. This approach is more typically used in  reinforcement learning, defining a system that learns a global meta-policy that supports efficient learning on specific tasks. The resulting RNNs have been shown to encode information equivalent to a Bayesian posterior distribution \citep{mikulik2020meta}, making them a good candidate for metalearning amortized Bayesian inference as we do here.

\section{Experiments}\label{sec:experiments}

We apply the neural circuit to two data settings: a synthetically generated dataset where the form of the DPMM is known, and a challenging open-set image classification task. The goal of our experiments is to compare both the predictive performance and computational efficiency of the neural circuit to standard sequential inference techniques for the DPMM.

\subsection{Evaluation Metrics}

We consider three main types of evaluation metrics. The first measures the quality of a model's predictions when the true class label is provided after each prediction. We refer to this as the \emph{sequential observation setting}. For this type of evaluation, we compute the average predictive negative log-likelihood (NLL) averaged across timesteps:
\begin{equation}
    \text{NLL}(\rvx_{1:T}, z_{1:T}) \triangleq -\frac{1}{T} \sum_{t=1}^T \log p(z_t | \rvx_{1:t}, z_{1:t-1}). \label{eq:nll_definition}
\end{equation}
We also report perplexity, which is simply the exponentiation of \eqref{eq:nll_definition}. If a model has perplexity level $v$, its probability of outputting the correct label can be loosely equated to the probability of obtaining a specific face upon rolling a $v$-sided die (note however that $v$ is not necessarily an integer). Therefore lower perplexity is better and a perplexity of 1 is the best. The second type of metric we consider quantifies how well the model predicts when there is no feedback about the true class labels (we refer to this as the \emph{fully unobserved setting}). We compare the maximum a posteriori (MAP) prediction of the model against the true sequence of class labels using a clustering metric, such as adjusted Rand index (ARI)~\citep{hubert1985comparing} or adjusted mutual information score (AMI)~\citep{vinh2009information}. Finally, in order to compare computational efficiency, we evaluate the wall clock time per sequence to make predictions both in the sequential observation and fully unobserved settings.

\subsection{Experimental Details}
As full batch approaches to DPMM inference such as Gibbs sampling do not scale well to the sequential inference setting, our main experimental point of comparison is the particle filter of \citet{fearnhead2004particle} discussed in Section~\ref{sec:background_particle_filter}.

For the particle filter, we use 100 particles for inference and utilize an adaptive resampler that resamples according to a multinomial distribution over the particles whenever the effective sample size drops below 50. In the fully unobserved setting, we take the MAP prediction to be the particle with the largest weight after running the filter on a sequence.

For the neural circuit, we use a 2-layer gated recurrent unit (GRU)~\citep{cho2014properties} network with 1024 hidden units. We metalearn the neural circuit by minimizing the NLL of sequences sampled from the training sequence distribution. We train the neural circuit using Adam~\citep{kingma2015adam} with learning rate $0.001$ for 10,000 gradient descent steps using minibatches of 128 sequences of length $T=100$. MAP prediction is implemented in the neural circuit by simply selecting the class label with highest predicted probability for each timestep.

All experimental results are evaluated as the average over 10,000 held-out sequences of length $T=100$. All methods are implemented in PyTorch~\citep{paszke2019pytorch} and are GPU-enabled.  Our code is publicly available on Github\footnote{\url{https://github.com/jakesnell/neural-circuits}}. For additional experimental details, please refer to Appendix~\ref{sec:experimental_details}. 

\subsection{Modeling Synthetic Data from a DPMM}\label{sec:synthetic_dpmm_results}
\begin{figure*}[ht]
    \centering
    \includegraphics[trim=1em 35em 52em 0em, clip, width=\textwidth]{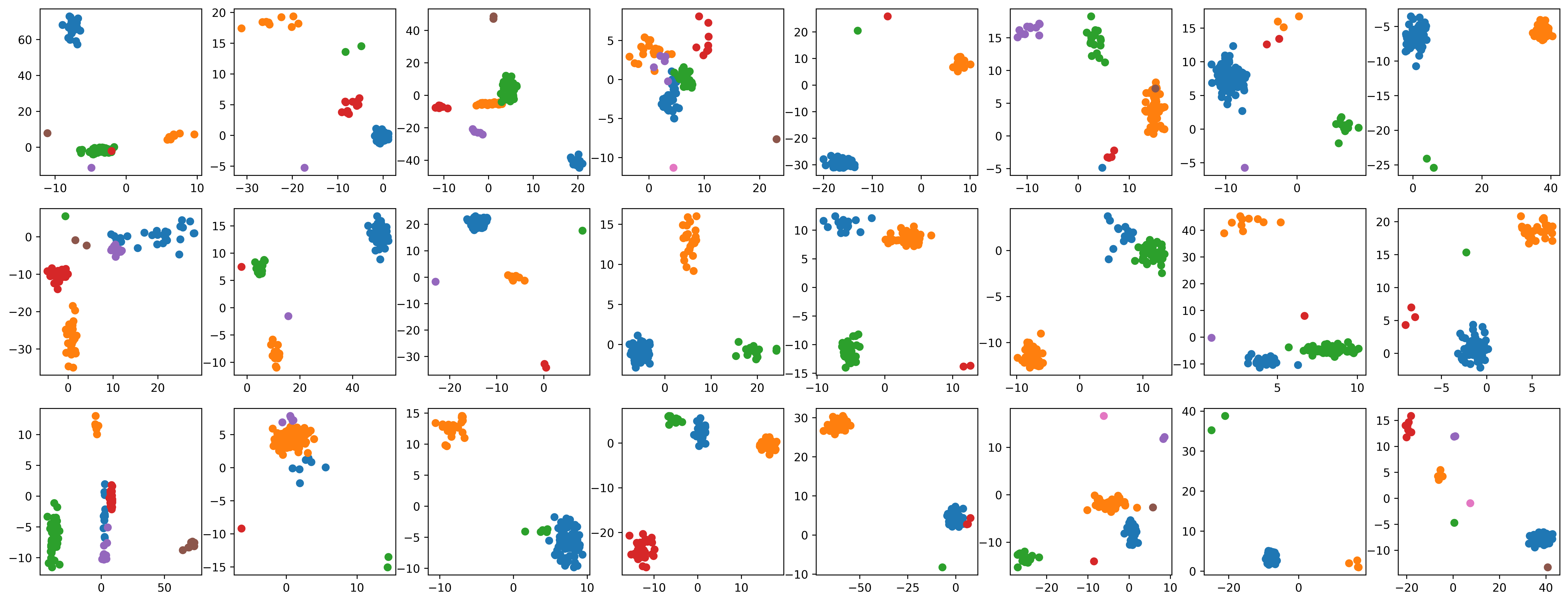}
    \caption{Visualization of sample synthetic sequences generated from the normal-inverse-prior used in Section~\ref{sec:synthetic_dpmm_results}. Classes are sampled from a Chinese restaurant process (CRP) with $\alpha = 1.0$ and sequences consist of 100 timesteps. Clusters are colored by the true class label.}
    \label{fig:synthetic_visualization}
\end{figure*}

We first evaluate performance of the neural circuit on a synthetically generated DPMM dataset. Our aim is to determine whether the neural circuit can match performance of the particle filter when the class-conditional distributions belong to the exponential family with a known prior. Specifically, we use a normal-inverse-gamma prior and Gaussian class conditional distributions with unknown mean and variance. For each dimension $d = 1, \ldots, D$, the form of the class-conditional distribution is:
\begin{align}
    \sigma_d^2 &\sim \Gamma^{-1}(a, b) \label{eq:synthetic_distribution} \\
    \mu_d &\sim \mathcal{N}(m, \sigma_d^2 / \lambda) \nonumber \\
    x_d &\sim \mathcal{N}(\mu_d, \sigma_d^2), \nonumber
\end{align}
where $m$, $\lambda$, $a$, and $b$ are known hyperparameters. We set the length of the sequences to $T=100$ and chose $D = 2$, $m = 0$, $\lambda = 0.01$, and $\alpha = \beta = 2$. A visualization of several sequences drawn from this distribution is shown in Figure~\ref{fig:synthetic_visualization}. We directly set the hyperparameters of the particle filter to their true values.

The results of our evaluation can be found in Table~\ref{tab:synthetic_dpmm_results}. We find that although the particle filter performs best on negative log-likelihood, the neural circuit provides better clusterings in the fully unobserved setting. We hypothesize this is due to the particle filter's insufficient exploration of the posterior over labelings stemming from its reliance on particles. We also find that the neural circuit inference is roughly $5\times$ faster than the particle filter in the sequential observation setting and approximately $10\times$ faster in the fully unobserved setting.
\begin{table*}[ht]
    \centering
    \caption{Results on two-dimensional data synthesized from a DPMM. Evaluation computed as average over 10,000 held-out sequences of length 100. Negative log-likelihoods are expressed in nats per timstep.}
    \begin{tabular}{lcccccc}
    \toprule
    &\multicolumn{2}{c}{Sequential Observation}&\multicolumn{2}{c}{Fully Unobserved}&\multicolumn{2}{c}{Inference Time (ms/sequence)} \\
    \cmidrule(lr){2-3}\cmidrule(lr){4-5}\cmidrule(lr){6-7}
    Method & NLL (\textdownarrow) & Perplexity (\textdownarrow) & ARI (\textuparrow) & AMI (\textuparrow) & Seq. Obs. (\textdownarrow) & Fully Unobs. (\textdownarrow) \\ \midrule
    CRP & 1.0055 & 2.9782  & 0.0101  & 0.0101 & \textbf{0.0192} & \textbf{0.1792} \\
    Particle Filter & \textbf{0.0484} & \textbf{1.0528} & 0.7691 & 0.8144 & 1.6168 & 4.4321 \\
    Neural Circuit & 0.0746 & 1.0847 & \textbf{0.9225} & \textbf{0.9293} & 0.3153 & 0.4515\\
    \bottomrule
    \end{tabular}
    \label{tab:synthetic_dpmm_results}
\end{table*}
\subsection{Open-set Classification on ImageNet-CRP}
Next we consider a challenging open-set image classification task where the input features are activations from a ResNet~\citep{he2016deep}. Our goal is to determine whether the neural circuit is able to effectively scale up to a high dimensional space where the form of the class-conditional distribution is no longer known.

We downloaded the weights of a pretrained ResNet-18 from TIMM~\citep{rw2019timm} and extract 512-dimensional penultimate layer activations from the entire ILSVRC 2012 dataset~\citep{russakovsky2015imagenet}. We split the 1,000 classes into 500 reserved for training (meta-train classes) and 500 for testing (meta-test classes). We generate sequences by first sampling $z_{1:N}$ from a CRP with $\alpha = 1.0$. We assign each distinct value of $z_n$ to a class uniformly at random and then sample uniformly without replacement from the images belonging to that class in order to generate the corresponding observation $x_n$. We call this data-generating procedure ImageNet-CRP and note that it bears similarities to long-tailed datasets \citep{cui2019class,vanhorn2018inaturalist} except that across sequences different classes can play the role of the dominant or long-tailed classes.

We metalearn the neural circuit with the same architecture and optimization procedure as in Section~\ref{sec:synthetic_dpmm_results}. For the particle filter, additional care needs to be taken to model the sparse nonnegative ResNet features that are produced by a ReLU activation. We modeled this using a hurdle~\citep{cragg1971several} model with log-normal distribution over nonegative values. This model posits a log-normal distribution over the nonnegative values and a point mass at zero:
\begin{align}
    x_d &\sim \tilde{x}_d \cdot y_d \\
    \tilde{x}_d &\sim \LogNormal(\mu_d, \sigma_d^2) \nonumber \\
    y_d &\sim \Bern(p_d) \nonumber
\end{align}
It can be shown (see Appendix~\ref{sec:hurdle_derivation} for details) that $x_d$ can be expressed as being drawn from exponential family with a Beta prior on the hurdle probability $p_d$ and a normal-inverse-gamma prior on the log-normal parameters $\mu_d$ and $\sigma_d^2$. We apply this exponential family model independently to each of the 512 dimensions. Since the optimal hyperparameters for the particle filter are \emph{a priori} unknown, we metalearn the hyperparameters of the conjugate prior using the same minibatch setup as the neural circuit and Adam with learning rate of $0.1$. Note that gradient estimation with respect to the hyperparameters is possible in the particle filter since the NLL of a sequence with sequential observations can be computed without the use of particles. 

The results of this experiment are show in Table~\ref{tab:resnet_dpmm_results}. Despite the effort to adapt the particle filter to this setting by carefully selecting the exponential family model, the neural circuit outperforms the particle filter by a large margin, both in terms of predictive performance and computational efficiency. Here, fully unobserved inference in the neural circuit is over $100\times$ faster than the particle filter, since exponential family inference in the particle filter must be performed separately over each of the 512 dimensions. As expected, predictive performance of the neural circuit drops when evaluating on novel classes drawn from the meta-test set, as these novel classes represent patterns of activations the circuit has not encountered during metalearning. However, the neural circuit still significantly outperforms the particle filter even in this difficult setting. Importantly, the architecture and training procedure of the neural circuit is identical to our setup in the experiment with synthetic data (Section~\ref{sec:synthetic_dpmm_results}), which speaks to the versatility of our method.

\begin{figure*}
  \centering
    \includegraphics[width=400px]{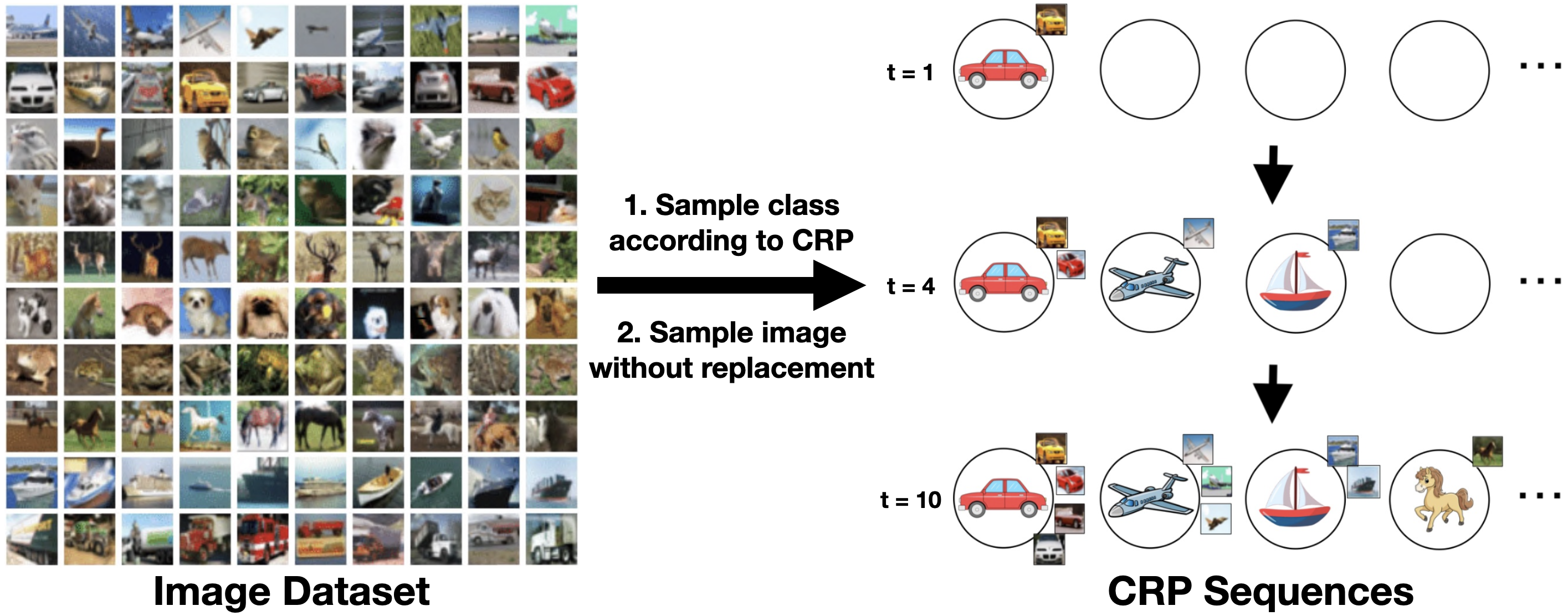}
    \caption{Diagrammatic representation of generating a Chinese restaurant process (CRP) using image data. At every sequence step $t$, a class is sampled according to a CRP and then an image from that class is sampled without replacement.} %
    \label{fig:cifar_crp}
\end{figure*}

\begin{table*}[ht]
    \centering
    \caption{Results on Imagenet-CRP with ResNet-18 activations as features. Evaluation computed as average over 10,000 held-out sequences of length 100. Negative log-likelihoods are expressed in nats per timstep.}
    \begin{tabular}{lcccccc}
    \toprule
    &\multicolumn{2}{c}{Meta-train Classes}&\multicolumn{2}{c}{Meta-test Classes}&\multicolumn{2}{c}{Inference Time (ms/sequence)} \\
    \cmidrule(lr){2-3}\cmidrule(lr){4-5}\cmidrule(lr){6-7}
    Method & NLL (\textdownarrow) & ARI (\textuparrow) & NLL (\textdownarrow) & ARI (\textuparrow) & Seq. Obs. (\textdownarrow) & Fully Unobs. (\textdownarrow) \\ \midrule
    CRP & 1.0052 & 0.0099 & 1.0032 & 0.0093 & \textbf{0.0190} & \textbf{0.1852} \\
    Particle Filter & 0.8478 & 0.0699 & 0.9333 & 0.0481 & 2.4073 & 73.8964 \\
    Neural Circuit & \textbf{0.2331} & \textbf{0.7720} & \textbf{0.6639} & \textbf{0.2802} & 0.3642 & 0.5013 \\
    \bottomrule
    \end{tabular}
    \label{tab:resnet_dpmm_results}
\end{table*}

\section{Related Work}

Many nonparametric Bayesian models have approximate solutions available in the form of Markov chain Monte Carlo (MCMC) \citep{neal2000markov} or variational inference \citep{wainwright2008graphical}. Examples include Gaussian processes \citep{tran2016variational}, Indian-Buffet processes \citep{doshi2009variational, teh2007stick}, and the Dirichlet process models we consider here \citep{blei2006variational, rasmussen1999infinite}. These formulations, however, assume a batch setting that is not efficient for sequential inference. Nonparametric Bayesian inference can be made more efficient in sequential settings by considering online variational inference with fixed update costs \citep{campbell2021online} or various forms of variational particle filtering \citep{saeedi2017variational, naesseth2018variational, maddison2017filtering}. However, overly restrictive distributional assumptions (e.g.\ exponential family) can hinder performance when working with complicated feature spaces such as neural network representations.

Amortized inference \citep{kingma2014auto, rezende2014stochastic} is better suited to complicated problem domains since it learns a function that maps directly from inputs an approximate posterior distribution, effectively amortizing the cost of variational inference. Several works have considered amortized inference over Dirichlet priors \citep{nalisnick2017stick,ehsan2017infinite,joo2020dirichlet}, Gaussian process priors \citep{casale2018gaussian}, and nested Chinese Restaurant Process priors \citep{goyal2017nonparametric}. Similar to the classic approaches mentioned above, these methods belong to the family of batch inference solutions. Sequential variational autoencoders \citep{chung2015recurrent, fraccaro2016sequential, gregor2019temporal, marino2018general} offer an online solution but none of these approaches are suitable to performing sequential inference in a DPMM.

Our neural circuit approach offers a scalable solution to sequential nonparametric Bayesian inference in a DPMM. We do so by using metalearning to incorporate the inductive bias of nonparametric Bayes into the learned network rather explicitly instantiating such a model. This is reminiscent of previous metalearning approaches that train with ``episodes''~\citep{vinyals2016matching} to learn a general prior distribution over weight initializations \citep{finn2017model, grant2018recasting} or a prior distribution over languages \citep{mccoy2023modeling}. We use sequences of observations and labels that can also be viewed as a kind of episode, but our goal is different: we aim to perform sequential inference over an unbounded number of classes. Compared to the previously mentioned approaches using variational inference or amortized inference, we leverage metalearning to directly learn a distribution over class labels rather than jointly learning encoder and decoder networks to maximize a variational lower bound on the log likelihood of the observations.

Related to our aims, the goal of open set recognition (OSR) is, broadly speaking, to detect previously unseen classes while accurately predicting known classes \citep{scheirer2013open}. Several early approaches to OSR focus solely on detecting previously unseen classes, falling into those based on traditional machine learning methods \citep{cevikalp2016best,cevikalp2017fast,jain2014multi,scheirer2014probability,zhang2016sparse} and, more recently, those attempting to endow neural networks with open-set capabilities \citep{bendale2016towards,dhamija2018reducing,shu2017doc}.

\citet{bendale2015towards} first treated OSR as an incremental learning problem by using extracted image features to perform metric learning over known classes initially, and performing incremental class learning thereafter. This method employs thresholded distances from the nearest class mean as its decision function. \citet{rudd2017extreme} advanced this approach by introducing distributional information into the thresholding process. Both methods rely on large datasets of known classes. \cite{liu2020few} introduced a metalearning formulation of OSR based on thresholding prototypical embeddings \citep{snell2017prototypical} to address this limitation. Similarly, \citet{willes2022bayesian} proposed a method called FLOWR that combines prototypical embeddings with Bayesian nonparametric class priors. These methods operate in the sequential observation setting, in that they observe the true class label after every prediction, and explicitly learn a metric space end-to-end instead of performing inference over arbitrary input features (including possibly representations extracted from a pretrained network) as we do. The particle filter baselines~\citep{fearnhead2004particle} we compare against closely resemble a variant of FLOWR modified for our problem setting.

\section{Conclusion}

We have proposed neural circuits for sequential nonparametric Bayesian inference: metalearning a recurrent neural network to capture the inductive bias of a DPMM that generates the training sequences. Our approach has stronger predictive performance than particle filter baselines while also being more computationally efficient. Neural circuits are simple to implement yet flexible enough to handle complex inputs with minimal changes to their training procedure and architecture.  In future work, we plan to apply the neural circuit approach to capture the inductive bias of more complicated nonparametric Bayesian models with richer latent spaces.

\subsubsection*{Acknowledgements}
This work was supported by ONR (grant number N00014-23-1-2510). JCS gratefully acknowledges financial support from the Schmidt DataX Fund at Princeton University made possible through a major gift from the Schmidt Futures Foundation.

\bibliography{refs}

\appendix

\onecolumn
\section{Derivation of Hurdle Model}\label{sec:hurdle_derivation}

First we recall the form of an exponential family likelihood:
\begin{equation}
p(\rvx \mid \bm{\eta}) = h(\rvx) \exp \left\{ \langle \bm{\eta}, \rvt(\rvx) \rangle - A(\bm{\eta}) \right\}. \label{eq:exp_density}
\end{equation}
The corresponding conjugate prior takes the following form:
\begin{equation}
p(\bm{\eta} \mid \bm{\tau}, \nu) = \exp \left\{ \langle \bm{\tau}, \bm{\eta} \rangle - \nu A(\bm{\eta}) - B(\bm{\tau}, \nu) \right\}. \label{eq:conjugate_prior}
\end{equation}

\subsection{Beta-Bernoulli}\label{sec:beta_bernoulli}

We first show how a Bernoulli likelihood and Beta conjugate prior can be expressed in terms of \eqref{eq:exp_density} and \eqref{eq:conjugate_prior}. This is a useful stepping stone towards the hurdle model, which contains the Bernoulli distribution as a component. Recall that for a Bernoulli likelihood $\Bern(x|\theta)$,
\begin{align}
    p(x | \theta) &= \theta^x (1-\theta)^{1-x} \\
    \log p(x | \theta) &= x \log \theta + (1 - x) \log (1-\theta),
\end{align}
from which we recognize
\begin{align}
    \eta &= \log \frac{\theta}{1 - \theta} \label{eq:bernoulli_natural} \\
    t(x) &= x \nonumber \\
    A(\eta) &= -\log(1 - \theta). \nonumber
\end{align}
Similarly, recall that for a Beta conjugate prior $\Beta(\theta | a, b)$, 
\begin{align}
    p(\theta | a, b) &= \frac{\Gamma(a + b)}{\Gamma(a) \Gamma(b)} \theta^{a-1} (1 - \theta)^{b - 1} \\
    \tau &= a - 1 \nonumber \\
    \nu &= a + b - 2 \nonumber \\
    B(\tau, \nu) &= \log \Gamma(a) + \log \Gamma(b) - \log \Gamma(a + b). \nonumber
\end{align}

\subsection{Hurdle Model}
Now suppose we have an arbitrary exponential family likelihood $p(x | \bm{\eta})$ of the form \eqref{eq:exp_density} and a conjugate prior $p(\bm{\eta} | \bm{\tau}, \nu)$ of the form  \eqref{eq:conjugate_prior}. Now define a hurdle model based on this likelihood to be:
\begin{equation}
    q(x | \tilde{\bm{\eta}}) = \left \{ 
    \begin{array}{cl}
    1 - \theta & \mbox{, $x = 0$} \\
    \theta p(x | \bm{\eta}) & \mbox{, $x \neq 0$}.
    \end{array}
    \right .
\end{equation}
Intuitively, this is an application of a Bernoulli variable that gates whether the underlying exponential family likelihood is active or not. Observe that $q(x | \tilde{\bm{\eta}})$ may be written as:
\begin{align}
    q(x | \tilde{\bm{\eta}}) &= (1 - \theta)^{(1 - \mathbbm{1}\{ x \neq 0 \})} (\theta h(x) \exp \left \{ \langle \bm{\eta}, \rvt(x)  \rangle - A(\bm{\eta})\right\})^{\mathbbm{1} \{ x \neq 0 \} } \\
    \log q(x | \tilde{\bm{\eta}}) &= \log (1-\theta) + \mathbbm{1} \left\{ x \neq 0 \right \} h(x) + \langle \bm{\eta}, \rvt(x) \mathbbm{1} \left \{ x \neq 0 \right \} \rangle + \left( \log \frac { \theta } { 1 - \theta } - A(\bm{\eta}) \right) \mathbbm{1} \left \{ x \neq 0 \right \}
\end{align}
We can therefore recognize $q(x | \tilde{\bm{\eta}})$ itself as exponential family:
\begin{align}
q(x | \tilde{\bm{\eta}}) &= \tilde{h}(x) \exp \{ \langle \tilde{\bm{\eta}}, \tilde{\rvt}(x) \rangle - \tilde{A}(\tilde{\bm{\eta}}) \} \\
\tilde{\bm{\eta}} &= \left[ \begin{array}{c} \log \frac{\theta}{1-\theta} - A(\bm{\eta}) \\ \eta_1 \\ \eta_2 \\ \vdots \end{array}\right] \nonumber \\ 
\tilde{\rvt}(x) &= \left[ \begin{array}{c} \mathbbm{1} \{ x \neq 0 \} \\ t_1(x) \mathbbm{1} \{ x \neq 0 \} \\ t_2(x) \mathbbm{1} \{ x \neq 0 \} \\ \vdots \end{array} \right] \nonumber \\
\log \tilde{h}(x) &= \mathbbm{1} \{ x \neq 0 \} \log h(x) \nonumber \\
\tilde{A}(\tilde{\bm{\eta}}) &= -\log(1-\theta) \nonumber
\end{align}
We also posit a conjugate prior for $q(x | \tilde{\bm{\eta}})$:
\begin{align}
q(\tilde{\bm{\eta}} | \tilde{\bm{\tau}}, \tilde{\nu}) &= \exp \{ \langle \tilde{\bm{\tau}}, \tilde{\nu} \rangle - \tilde{\nu} \tilde{A}(\tilde{\bm{\eta}}) - \tilde{B}(\tilde{\bm{\tau}}, \tilde{\nu}) \} \\
\log q(\tilde{\bm{\eta}} | \tilde{\tau}, \tilde{\nu}) &= \tilde{\tau}_1 \tilde{\eta}_1 + \langle \bm{\tau}, \bm{\eta} \rangle - \tilde{\nu} \tilde{A}(\tilde{\bm{\eta}}) - \tilde{B}(\tilde{\bm{\tau}}, \tilde{\nu})
\nonumber 
\end{align}
Now let $\Bern(x | \theta) = \bar{h}(x) \exp \{ \langle \bar{\eta}, \bar{t}(x) \rangle - \bar{A}(\bar{\eta}) \}$, with definitions following from \eqref{eq:bernoulli_natural}. In this case, $\bar{\eta} = \log \frac{\theta}{1 - \theta}$ and thus $\tilde{A}(\tilde{\bm{\eta}}) = \bar{A}(\bar{\eta})$. Moreover, $\tilde{\eta}_1 = \bar{\eta} - A(\eta)$, and we can then write
\begin{align}
    \log q(\bar{\eta}, \bm{\eta} | \tilde{\bm{\tau}}, \tilde{\nu}) &= \tilde{\tau}_1 ( \bar{\eta} - A (\bm{\eta})) + \langle \bm{\tau}, \bm{\eta} \rangle - \tilde{\nu} \bar{A}(\bar{\eta}) - \tilde{B}(\tilde{\tau}, \tilde{\nu}) \\
    &= \langle \bm{\tau}, \bm{\eta} \rangle - \tilde{\tau}_1 A(\bm{\eta}) + \tilde{\tau}_1 \bar{\eta} - \tilde{\nu} \bar{A}(\bar{\eta}) - \tilde{B}(\tilde{\tau}, \tilde{\nu}), \nonumber
\end{align}
from which we recognize that
\begin{align}
    \tilde{B}(\tilde{\tau}, \tilde{\nu}) &= B(\tau, \nu = \tilde{\tau}_1) + \bar{B}(\tilde{\tau_1}, \tilde{\nu}).
\end{align}
Therefore, the hurdle applied on top of an arbitrary exponential family likelihood can be expressed in terms of the underlying base model and the Beta-Bernoulli presented in Section~\ref{sec:beta_bernoulli}.

\section{Experimental Details}\label{sec:experimental_details}

The neural circuit RNN cell was chosen to be a 2-layer GRU with hidden size 1024. The number of output logits was set to 100. Training was performed over 10,000 minibatches each of size 128 sequences. The CRP coefficient was set to $\alpha = 1.0$ in all cases. The neural circuit was trained using Adam with learning rate $0.001$ and the particle filter using learning rate $0.1$. Tuning hyperparameters (e.g. model architecture, learning rates) were selected by monitoring training loss over the first several hundred minibatches and modifying as necessary. As a vast number of sequences can be generated, overfitting is much less of a concern in our setting compared to standard applications of deep learning. 

One random seed was used for training each method. Evaluation was done on 10,000 held-out sequences each of length 100. The clustering metrics were computed using the \texttt{adjusted\_rand\_score} and \texttt{adjusted\_mutual\_info\_score} functions from Scikit Learn~\citep{scikit-learn}. Experiments were performed on an internal cluster using NVIDIA A100 GPU nodes with 80 GB of GPU memory.

\end{document}